\documentclass[11pt]{article}

\usepackage[utf8]{inputenc}
\usepackage[T1]{fontenc}
\usepackage{lmodern}
\usepackage[margin=1.05in]{geometry}
\usepackage{booktabs}
\usepackage{graphicx}
\usepackage{amsmath}
\usepackage{caption}
\usepackage{microtype}
\usepackage[hidelinks,breaklinks]{hyperref}
\usepackage{url}

\newcommand{\KO}{\textsf{Known Other}}
\newcommand{\CBL}{\textsf{Unknown but could be lead}}

\title{\bfseries Auditing Recorded Predictive Lead Service-Line\\
Classifications Against Physical Verification:\\
A Statewide Study of New York}

\author{Muhammad Sarmad Sohail\thanks{ORCID 0009-0008-2776-5253.
Data, code and reproduction instructions are in Section \ref{sec:repro}.}\\
\normalsize Independent researcher}

\date{}

\begin{document}
\maketitle

\begin{abstract}
\noindent
Under the US Lead and Copper Rule Revisions, a water utility may determine a service line's
material with a statistical or predictive model instead of physically inspecting it. New
York State publishes, per address, which method was used, which makes those model outputs
auditable against physical verification performed by the same utilities in the same places.
Almost no address carries both, so throughout this paper the comparison is between two
populations inside one utility rather than between a prediction and an excavation at the same
address.

We screen all 153 New York localities, the finest reporting unit the file publishes, that
classified at least 100 addresses this way. \textbf{Seventy-five of them (49\%), covering
125,990 addresses or 57\% of the model-classified lines screened, record exactly one distinct
material value.} Zero variance alone is not misconduct, and we do not treat it as such: 68 of
the 75 are consistent with their own physical verification, or have too little of it to check.
\textbf{Seven are contradicted by their own crews}, six of them beyond any sampling
explanation. Five are the boroughs of New York City, which file as one water system; one is
East Rochester, an unrelated utility 550\,km away that reproduces the pattern exactly.

New York City is the largest case. A predictive model is the recorded basis for 43,215
addresses, and on all of them the recorded material takes a single value: \KO{}, the state
template's term for a line whose lead status is known to be negative but whose material is
not named. The city has other values available and uses them heavily: \textsf{Unknown} on
121,779 addresses, 1,880 of which it had already excavated, and lead on 120,692. In the
model bucket both counts are zero, and from zero events in 43,215 the 95\% upper bound on the
rate is 0.0085\%. Across the rest of New York the same method records lead or the hedge \CBL{}
on 12.21\% of 176,888 addresses. We report that comparison with its two weaknesses attached:
93\% of the outright lead in it is a single city, Poughkeepsie, which has verified nothing
physically, and the hedge is a value New York City has never used under any method. The
within-city contrast depends on neither.

The model-cleared population is genuinely newer, with a median year built of 1984 against
1930, and a spatial join to NYC MapPLUTO shows that construction era accounts for about a
third of the gap and not the rest. Holding era fixed, records-based classification, also a
desk method and applied to 532,906 addresses, finds lead in every era, at rates from 4.3\% to
31.9\%. Physical verification finds it in every era, from 1.5\% to 14.5\%. The predictive
model finds it in none. Six era-aware estimators across three method families put the expected
number of lead public-side lines among the model-cleared addresses at roughly 1,150 to 1,450.

Two findings need no comparison population. New York's own guidance permits \KO{} where
records show post-1986 construction; 7,782 of these addresses are in pre-1940 buildings and
the installation-date field is empty on all of them. And the archived 2025 snapshot shows the
public-side determination did not exist then: it was created by copying a customer-side model
output, for a different pipe with a different owner.
\end{abstract}

\section{Introduction}\label{sec:intro}

The Lead and Copper Rule Revisions (LCRR) required every US community water system to publish
an inventory of its service lines by October 2024 \cite{lcrr2021}. Where a utility does not
know a line's material, the rule permits statistical or predictive methods in place of
excavation, and a commercial market grew around that permission. Utilities buy these models
because excavation costs orders of magnitude more per line.

Whether the models work is asked less often than it should be, for a structural reason: the
party best placed to validate a model is the vendor that sold it, and the utility that bought
it has little incentive to fund a check that could invalidate an inventory it has filed.

\paragraph{Scope of the claim.}
The inventory records what a utility filed, not what a model produced. We therefore name no
vendor, make no claim about any vendor's model, and report no accuracy figure. An output with
no variation is consistent with a model that is confident and wrong; with a workflow that
sends uncertain predictions to a crew and records the model only on the lines it cleared; and
with a reporting pipeline that discards uncertainty before filing. What we audit is the
published record, and whether it meets the conditions the state attaches to the values used in
it. Section~\ref{sec:nyc} returns to these three readings and tests them against each other
where the data allows. Two of our findings need no comparison group and no estimator, and they
are the ones we would defend first: the pre-1940 subgroup in Section~\ref{sec:nyc} and the
archived snapshot in Section~\ref{sec:data}.

\paragraph{Prior validation against excavation: Flint.}
One study exists. Goovaerts \cite{goovaerts2023} validated a 2017
geospatial model against 26,750 excavations covering roughly half of Flint's tax parcels and
reported an AUC of 0.9 for copper and galvanized material but 0.6 for lead, barely better
than chance, improving to 0.8 only after replacing the kriging method with a compositional
variant.

It does not close the question. Flint is one city, and the most intensively studied water
system in the world since 2016 \cite{chojnacki2017,abernethy2018}. The model validated there
predates the LCRR by seven years and is not one of the commercial products now sold into the
compliance market, and it was checked by the author of the model being checked. More
importantly for what follows, the failure mode we document is different in kind. In Flint the
model attempted ``lead'' sometimes and aimed poorly. Here, across half the water systems that
use one, the recorded output does not vary at all.

\paragraph{Contributions.}
\begin{enumerate}\itemsep2pt
\item A screen for zero output variance that runs as a single aggregation over a file New
      York already publishes, applied to all 153 localities that classified at least 100
      addresses with a model (Section~\ref{sec:screen}). We report it in two stages, because
      zero variance flags and only contradiction confirms.
\item An independent replication: East Rochester reproduces New York City's pattern in a
      different utility, region and size class.
\item A measurement of the largest case against the state's own written standard rather than
      against our judgement (Sections~\ref{sec:rules} and~\ref{sec:nyc}).
\item Separation of the part of the gap explained by the model being applied to newer housing
      stock from the part that is not, using construction era joined from NYC MapPLUTO, and
      eleven estimators of the residual with coverage and calibration reported
      (Section~\ref{sec:nyc}).
\item An open baseline from public covariates, with discrimination reported as a function of
      how far apart training and test data are held (Section~\ref{sec:baseline}).
\item Five parsing and semantic traps in the source file, one of which inflates New York
      City's apparent size twofold (Section~\ref{sec:data}).
\end{enumerate}

\section{What the regulation actually requires}\label{sec:rules}

New York's own guidance to water systems \cite{nysdohguidance2025} sets the standard against
which the rest of this paper measures. Three provisions matter.

\paragraph{A model output is not presumptively acceptable.}
On whether a predictive model may make a line ``known'' without physical verification, the
guidance states: \emph{``A model's output typically needs physical verification due to an
inherent inaccuracy of any model or statistical analysis. However, on a case-by-case basis,
some of the model and statistical analysis results will be accepted without physical
verification. You must provide sufficient information to the State to evaluate how much
physical verification is adequate.''} Among the information a system must supply, the
guidance lists \emph{``a random physical verification process such as the proposed number of
SLs that will be physically verified''} and \emph{``confidence interval for the model''}.

\paragraph{The rule contains a feedback loop.}
The guidance closes: \emph{``Note that a State's initial determination for a required physical
verification rate can be revised based on the accuracy of physical confirmation results.''}
The required verification rate is meant to respond to what the shovels find.

\paragraph{The vocabulary distinguishes ``not lead'' from ``might be lead''.}
Systems choose a material from eight values: \textsf{Lead including lead-lined galvanized},
\textsf{Copper}, \textsf{Galvanized}, \textsf{Plastic}, \KO{}, \CBL{}, \textsf{Unknown but
unlikely lead}, and \textsf{Unknown}. Three of the eight are explicit unknowns and one of
those three says the line could be lead. \KO{} sits with the named materials and maps the
line to non-lead in the state template; it asserts a known negative lead status without
naming a material. The guidance permits it in place of an actual material where the system
holds \emph{``written records showing the entire distribution system was constructed after
June 1986''} and the customer-owned portion was likewise built after the lead ban.

We are therefore not asking whether the model is accurate, a question this data cannot answer.
We are asking whether what was filed meets the standard the state wrote down.

\section{Data}\label{sec:data}

\paragraph{Source and snapshots.}
The New York State Lead Service Line Inventory \cite{nyslsl2026} is a public Socrata dataset.
We use two captures: \textbf{T1}, 2026-08-11, 4,618,115 rows, and \textbf{T0}, 2025-06-22,
3,747,025 rows, retrieved from the Internet Archive. Both are pinned by SHA-256 in
\cite{nyslsl2026}; the live dataset moves, so the checksum and not the endpoint is the
reference. Row counts parse exactly to the API's own counts.

\paragraph{Trap 1: rows are not lines.}
T1's 4,618,115 rows resolve to 3,744,223 distinct \mbox{locality}--\mbox{street}--\mbox{zip}
keys. The excess is concentrated: every New York City borough has a rows-per-address ratio of
2.001, while no other locality above 5,000 rows exceeds 1.14. New York City files two rows per
address, one carrying the public-side columns and one only the customer-side columns.

\paragraph{Trap 2: the public side was copied from the customer side, and we can date it.}
On the public-side row, the public verification method equals the customer method on 100.0\% of
rows, a perfect diagonal across all five methods, and the public material equals the customer
material on 100.0\%. The archived snapshot fixes the direction. At T0, New York City's 817,375
addresses appear once each, and \emph{the public columns are blank on every one of them}: its
43,440 model classifications exist only on the customer side. Fourteen months later the same
addresses carry the identical value on the public side as well. The public-side determination
was not made independently; it was created from a model output produced for the customer's
pipe, which under LCRR is a different pipe with a different owner.

\paragraph{Trap 3: read the material, not the category.}
The address-level category field collapses to four LCRR classes and, outside New York City,
reports the worse of the two sides: 9,625 addresses carry a public material that is
definitely not lead and a lead category driven by the customer side. Inside the city the two
coincide exactly, for the reason given in Trap 2. All results here read
\textsf{Current Public Side SL Material} directly.

\paragraph{Trap 4: the method and material fields are free text.}
Locality strings are case-split, and New York City's boroughs appear as the codes
\texttt{QN}, \texttt{BK}, \texttt{SI}, \texttt{BX}, \texttt{MN}: a literal filter on
``Queens'' returns 11 rows statewide. The model method appears as
\textsf{Statistical Analysis/Predictive Model} 219,899 times and under four further spellings
a further 4,603 times. \textbf{The screen in Section~\ref{sec:screen} matches the method
broadly}, since restricting to one spelling would be a selection the reader cannot check. The
rest-of-state comparison in Section~\ref{sec:nyc} uses the canonical spelling alone, and we
report what that choice costs: matching broadly moves the rate quoted there from 12.21\% of
176,888 addresses to 11.93\% of 181,065. None of the additional spellings occurs in New York
City, so no New York City figure depends on the choice.

\paragraph{Trap 5: the published column descriptions are shifted by one.}
In the dataset's own metadata, the description attached to
\textsf{Current Public Side SL Material} in fact describes the following column, and the
error propagates. The guidance document \cite{nysdohguidance2025}, not the portal metadata,
is the authority on field meaning.

\paragraph{Working population and external join.}
New York City contributes 817,375 addresses; we drop the 3,121 without a parseable coordinate
and the 343 (0.042\%) falling outside a generous bounding box for the city, leaving 813,911.
Of the model-classified addresses, 198 and 24 respectively are lost this way, and all 43,437
carry \KO{} before any exclusion, so no exclusion removes a lead value.
Construction date is 0.0\% populated in every New York City bucket, so we join each address to
its nearest MapPLUTO tax lot \cite{mappluto}: all 858,602 lots retrieved, 95.4\% with a usable
year built, median match distance 23.4\,m and 99.97\% inside a 100\,m cap.

\section{A statewide screen for zero output variance}\label{sec:screen}

\begin{figure}[t]\centering
\includegraphics[width=\textwidth]{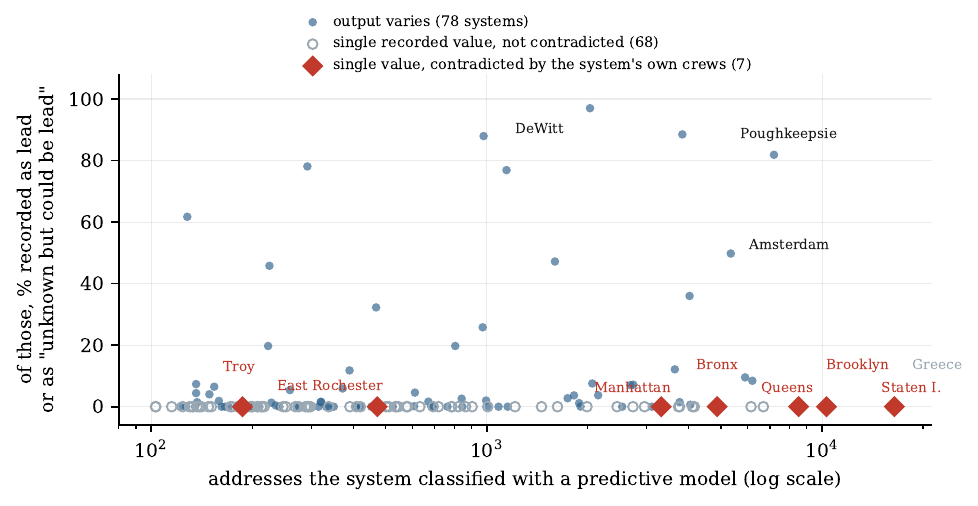}
\caption{Every New York water system with at least 100 model-classified addresses (153
systems, 221,090 addresses). Vertical axis is the share of those classifications recorded as
lead or as \CBL{}. Hollow markers on the floor are systems whose model output takes a single
value; red diamonds are the subset contradicted by the same system's own physical
verification. The field's full range is in live use: Poughkeepsie records lead on 81.9\% of
7,189 addresses, DeWitt hedges on 97.0\% of 2,033.}
\label{fig:screen}
\end{figure}

The screen is one aggregation: for each locality, how many distinct material values does its
predictive-model output take, and what does that locality's own physical verification find?

Of the 153 localities, \textbf{75 (49\%) record exactly one distinct value}, covering 125,990
addresses, \textbf{57.0\% of all model-classified lines screened}. That number alone proves
nothing. A system serving genuinely lead-free housing stock \emph{should} return one value,
and most of these do so defensibly: 68 of the 75 either have physical verification that also
finds essentially no lead, or have too little to test against. \textbf{Zero variance flags;
contradiction confirms.}

Seven localities fail the second stage, defined in advance as at least 200 physically verified
lines of which at least 1\% are lead (Table~\ref{tab:contra}). Five are New York City boroughs.

\paragraph{The unit is the locality, not the water system.}
\textsf{Service Line Locality} is the finest geography the file publishes, and we screen on
it. It is not the regulated entity: NYSDOH lists the five boroughs under one public water
system, New York City Water Supply System \mbox{NY7003493} \cite{nysdohsummary}. Collapsing
them gives 149 units, 71 with a single recorded value (48\%), covering the same 125,990
addresses, and \textbf{three contradicted rather than seven}. We report the locality figure
because it is the unit the data supports and the one a regulator would query, and we state the
system figure here so that neither is a surprise. Nothing below depends on which is used.

\begin{table}[t]\centering\small
\caption{Localities whose model output takes one value and whose own crews contradict it.
The last two columns ask how surprising the zero is: how many lead findings the model bucket
would contain if it shared the locality's own physically verified lead rate, and the
probability of observing none. Troy is the one case a referee should discount.}
\label{tab:contra}
\begin{tabular}{lrlrrrr}
\toprule
Locality & Model-classified & Only recorded value & Verified & Lead found
  & Expected & $P(0)$ \\
\midrule
Staten Island   & 16,434 & \KO{} & 20,500 & 2.31\%  & 380   & $10^{-167}$ \\
Brooklyn        & 10,308 & \KO{} & 52,024 & 7.54\%  & 777   & $10^{-351}$ \\
Queens          &  8,513 & \KO{} & 66,460 & 14.19\% & 1,208 & $10^{-566}$ \\
Bronx           &  4,867 & \KO{} & 17,769 & 5.46\%  & 266   & $10^{-119}$ \\
Manhattan       &  3,315 & \KO{} & 11,751 & 3.40\%  & 113   & $10^{-50}$ \\
East Rochester  &    472 & \KO{} &    557 & 9.69\%  & 46    & $10^{-21}$ \\
Troy            &    187 & \textsf{Copper} & 325 & 1.54\% & 3 & $0.055$ \\
\bottomrule
\end{tabular}
\end{table}

\paragraph{East Rochester is the result that matters most.}
It is a village of a few thousand households, 550\,km from New York City, with a different
utility and no plausible shared vendor relationship visible in the data. Its model classified
472 addresses and returned \KO{} on all of them, in a place where its own crews opened 557
lines and found lead in 9.69\% of them. All 557 were opened by excavation, not by the lighter
field inspection, so the contradiction does not rest on the weaker of the two physical
methods. New York City is not an anomaly of scale.

\paragraph{Troy.}
Its 187 model classifications sit against 325 verified lines carrying five lead findings, so
the expected count under its own rate is three and a zero is unsurprising ($P = 0.055$). We
leave it in the table because the screen's threshold was fixed in advance and removing a case
after seeing its p-value is the practice this paper criticises, but no argument here depends
on it: the other six span $10^{-21}$ to $10^{-566}$.

\paragraph{Two structural facts about checkability.}
The median system physically verifies 0.14 lines per model-classified line, and 23 of the 153
verify none at all, covering 40,468 model-classified addresses that no screen can test. Greece
is the sharpest instance: 16,135 model-classified addresses, a single recorded value, and 74
verified lines to check it against. We tested whether systems leaning harder on the model
verify less. Comparing the two as shares of a system's lines finds no relationship (Spearman
$-0.076$, $p = 0.35$). Comparing model share against verification \emph{per model-classified
line} does find one ($-0.433$, $p = 2 \times 10^{-8}$), but both forms share a term across the
two variables and so carry a ratio artefact; we draw no conclusion from either beyond the
level being low throughout.

\paragraph{The feedback loop the guidance describes is not closing.}
Between T0 and T1, on an address basis, the model-classified population moved from 298,641 to
298,704 and the number of those classifications naming lead or the hedge \CBL{} moved from
28,971 to 28,974: 63 addresses and three findings in fourteen months, while physical
verification statewide
ran to 664,259 lines. The verification rate is meant to be revisable in light of what the
shovels find \cite{nysdohguidance2025}. In this file the model-classified population is
effectively frozen.

\section{New York City in detail}\label{sec:nyc}

\begin{table}[t]\centering\small
\caption{Public-side classifications in New York City by recorded basis, 813,911 addresses.
``Unknown'' counts materials recorded as some form of unknown; median year built from the
MapPLUTO join.}
\label{tab:methods}
\begin{tabular}{lrrrrr}
\toprule
Basis of classification & Addresses & Share & Lead & Unknown & Median built \\
\midrule
Records                   & 532,906 & 65.5\% & 19.80\% & 9.26\%   & 1931 \\
Field inspection          &  95,087 & 11.7\% &  8.45\% & 0.01\%   & 1931 \\
Excavation                &  72,174 &  8.9\% &  9.86\% & 2.60\%   & 1930 \\
Not verified              &  70,529 &  8.7\% &  0.00\% & 100.00\% & 1930 \\
\textbf{Predictive model} & \textbf{43,215} & \textbf{5.3\%} & \textbf{0.00\%} & \textbf{0.00\%} & \textbf{1984} \\
\bottomrule
\end{tabular}
\end{table}

Two rows in Table~\ref{tab:methods} return no lead. \textsf{Not verified} returns none because
it records nothing: all of its addresses are unknown. The predictive model returns none while
recording a determination on every address. Its 43,215 classifications take one value, \KO{}.

\paragraph{The within-city comparison.}
The strongest form of the finding needs no other utility. New York City has a value for a line
whose lead status is not established, \textsf{Unknown}, and it is in heavy use: 121,779
addresses, of which 49,365 were classified from records and 1,880 had already been excavated.
The city records lead on 120,692 addresses. Both counts are zero in the model bucket, on all
43,215 of it. These are the same filer, the same template and the same reporting period, and
every other basis of classification in Table~\ref{tab:methods} produces both values. From zero
events in 43,215 the 95\% upper bound on the rate is 0.0085\%.

\paragraph{Comparison with the rest of the state.}
Elsewhere in the state the same method uses more of the vocabulary: across 176,888 addresses
outside the city it records lead on 3.57\%, the hedge \CBL{} on a further 8.64\%, 12.21\%
together, and also \textsf{Copper}, \textsf{Plastic}, \textsf{Galvanized} and
\textsf{Unknown but unlikely lead}. We give the number because it is the natural question, and
then two reasons not to lean on it. First, the lead component is one city: 5,885 of the 6,320
rest-of-state model lead findings, 93\%, are Poughkeepsie, which has physically verified
nothing at all, and excluding it the outright rate falls to 0.26\% and the combined rate to
9.26\%. Poughkeepsie may be over-calling by as much as we argue New York City is under-calling.
We cite it to show that the field's full range is in live use, and we do not treat its rate as
a benchmark. Second, the
hedge is not a value the city has ever produced. Its public-side vocabulary is four values
across all 817,375 addresses and all five methods (\KO{}, \textsf{Lead including
lead-lined galvanized}, \textsf{Unknown} and \textsf{Galvanized}), with \CBL{},
\textsf{Unknown but unlikely lead}, \textsf{Copper} and \textsf{Plastic} used zero times under
any method. So 8.64 of the 12.21 points is a value the filer never emits anywhere, and only
the within-city contrast above is free of that.

\paragraph{What this does not attribute.}
The three readings set out in Section~\ref{sec:intro} apply here: a model that is confident and
wrong, a workflow that routes uncertain predictions to a crew and records the model only on
lines it cleared, and a pipeline that discards uncertainty before filing. The second is a
defensible way to run a utility. To every reader of the published inventory it is also
indistinguishable from the first, and the guidance asks for a confidence interval
\cite{nysdohguidance2025} precisely so that it would not be.

\paragraph{The pipeline explanation.}
The third reading deserves stating at its strongest, because it is the one a filer would
offer. New York City's pipeline is effectively binary, so any determination that is not lead
and not unknown collapses to \KO{}, and the four-value vocabulary above is evidence of exactly
that; the model may well have produced a distribution that the pipeline then flattened. We
accept that much, and it is why this paper claims nothing about the model's internals. What it
does not account for is the counts above. A binary pipeline still leaves \textsf{Unknown} available, and the city uses it 121,779
times and never once in the model bucket. \textsf{Records}, the other desk method with no site
visit, works inside the same four values in the same city and returns lead 19.80\% of the
time. A narrow vocabulary explains why the model bucket cannot say \textsf{Copper}. It does
not explain why that bucket alone never says \textsf{Unknown} or \textsf{Lead}.

\subsection{Construction era explains part of the gap}

The model was not applied at random: its addresses are newer, median year built 1984 against
1930, with 77\% in post-1960 buildings against 24\% of the physically verified population.
New York City did not ban lead pipes until 1961 and the federal ban came in 1986
\cite{osc2026}, so this alone predicts a lower true rate, and any comparison ignoring it
overstates the finding.

\begin{figure}[t]\centering
\includegraphics[width=\textwidth]{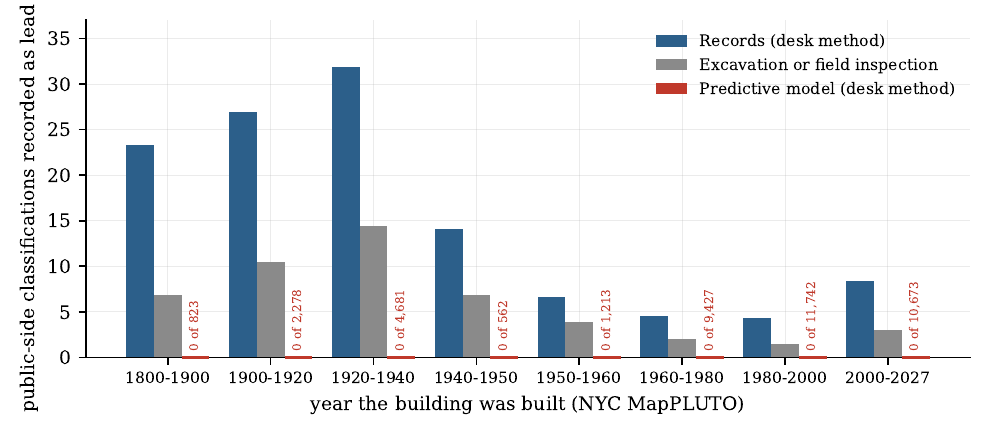}
\caption{Lead rate by construction era and basis of classification, New York City.
Records-based classification is a desk method and finds lead in every era; physical
verification finds lead in every era; the predictive model finds it in none, including 4,681
addresses in buildings built 1920--1940 where every other method in the same city finds lead
at 14--32\%.}
\label{fig:era}
\end{figure}

Figure~\ref{fig:era} holds era fixed. Records-based classification, also a desk method with no
site visit and the largest bucket in the city at 532,906 addresses, finds lead at 19.80\%
overall and between 4.32\% and 31.85\% in every era. Physical verification finds between
1.45\% and 14.45\%. The predictive model finds none in any era. ``It is a desk method, so of
course it is cautious'' does not survive the figure: the other desk method is the one that
finds the most lead.

\paragraph{The subgroup that requires no estimator.}
7,782 model-cleared addresses (18.0\%) sit in pre-1940 buildings: Brooklyn 2,679, Manhattan
1,991, Queens 1,572, the Bronx 861, Staten Island 679. The state's own grant programme uses
the count of pre-1939 homes as its lead-risk proxy \cite{osc2026}, and physical verification
of pre-1940 New York City buildings finds lead at 12.93\% (95\% CI 12.72--13.14\%,
$n = 98{,}749$). That rate applied to that subgroup implies roughly 1,000 lead lines recorded
as \KO{} in pre-1940 buildings alone. \textbf{The guidance permits \KO{} in place of an actual
material where records show post-1986 construction \cite{nysdohguidance2025}. These buildings
predate that by half a century, and the installation-or-replacement-date field is empty on
100\% of them}, so the published inventory contains no record of the kind the guidance
contemplates. The guidance also says what to do in that case, in the same item: \emph{``If you
do not have such records, you need to verify service line material with one or more methods
included in Item 14''}, that is, by one of the physical methods. The fallback is triggered by
the guidance's own terms. We note that absence of a date in the filing is not proof that the
utility holds no records; it is proof that the inventory as published does not show them.

\subsection{Estimating the residual}

\paragraph{Estimand.}
Essentially no address is both predicted and verified: 66 address keys statewide and one in
New York City carry both a model row and a physical row, far too few to identify an accuracy
figure, and we report none. Let $s$ index a stratum of borough, building type,
construction era and spatial neighbourhood, and $\hat\pi(s)$ the observed lead rate among
physically verified addresses in $s$. Under conditional exchangeability, that within $s$ the
choice of which addresses were dug is not further related to true material, $\hat\pi(s)$
estimates the counterfactual rate for the model-cleared addresses in $s$.

This is an assumption, not a randomisation. Its residual bias has a nameable direction: a
utility that trusts a model enough to skip the shovel plausibly leaves easier-looking cases to
the model and sends crews where something already looks wrong, which biases $\hat\pi(s)$
upward. What we can report is how far the estimate moves when the largest observable
confounder is added.

\begin{figure}[t]\centering
\includegraphics[width=\textwidth]{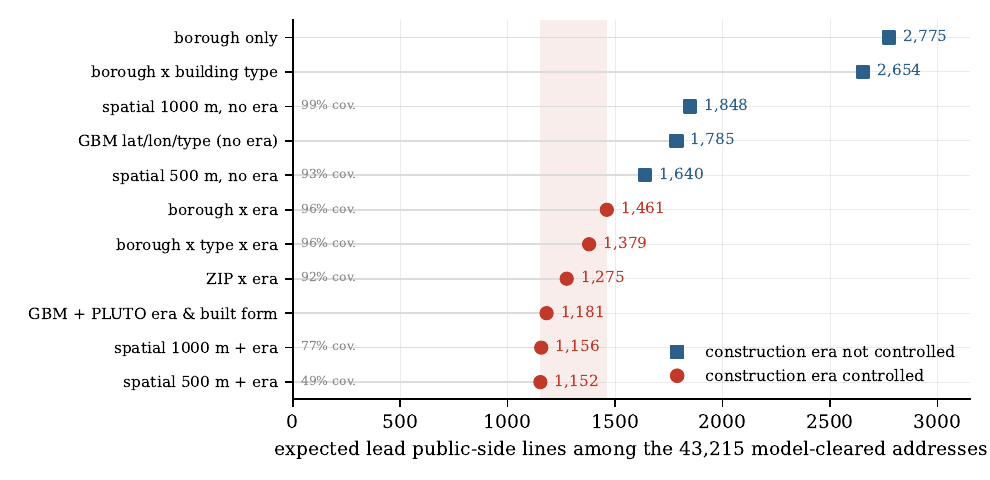}
\caption{Eleven estimators in three families: direct stratification, non-parametric spatial
nearest-neighbour matching at two calipers, and gradient-boosted trees. Estimators controlling
for construction era are red; the shaded band spans them. Where a stratum has fewer than 20
verified neighbours its addresses are counted unscored and the total extrapolated over the
covered share rather than dropped; coverage is annotated below 100\%.}
\label{fig:est}
\end{figure}

The five era-blind estimators span 1,640--2,775; the six era-aware ones span 1,152--1,461, or
2.7--3.4\% of the model-cleared population. Controlling for era removes about a third, in the
direction the bias argument predicted. \textbf{We report the era-aware band: on the order of
1,150--1,450 lead public-side lines among the 43,215 addresses recorded as \KO{}}, five
boroughs, public side only.

\paragraph{Uncertainty in the band.}
It is a spread across specifications, not a confidence interval, and the distinction matters
because the two are far apart in size. A nonparametric bootstrap over the physically verified
sample, 400 draws, puts sampling error on any single estimator at roughly $\pm 3\%$: borough
$\times$ era gives 1,461 with a 95\% interval of 1,419 to 1,503, borough $\times$ type
$\times$ era 1,379 (1,331 to 1,421), and ZIP $\times$ era 1,275 (1,227 to 1,330). The
1,152--1,461 range is therefore almost entirely disagreement between specifications rather
than sampling noise. Neither quantity says anything about the identifying assumption, which is
the larger source of doubt and is not a statistical one.

The two estimators that set the floor score only 49\% and 77\% of the population and
extrapolate over the covered share, and that extrapolation is not coverage-neutral. The
addresses the 500\,m spatial matcher can score are markedly older than those it cannot: 16.1\%
of the scored sit in 1920--40 buildings against 6.2\% of the unscored, and the corresponding
stratified lead rates are 3.5\% and 2.9\%. Dividing by coverage projects a higher-risk subset
onto a lower-risk remainder, so that estimator's 1,152 is high by roughly 10\% on this account.
The reader can take this either way and we do not choose: reweighting the two low-coverage
estimators moves the floor down to about 1,050, while dropping them and keeping only the four
above 90\% coverage moves it up to 1,181. Both leave the ceiling at 1,461, and neither changes
anything the paper concludes.

\paragraph{The convergence is weaker evidence than it looks.}
The three families use the same features on the same data under the same identifying
assumption. Their agreement shows the estimate is not an artefact of one functional form and
nothing more. The screen in Section~\ref{sec:screen} and the two dated facts in
Section~\ref{sec:data} rest on none of it.

\section{An open baseline, and what its accuracy bounds}\label{sec:baseline}

We fit a gradient-boosted tree on the 165,063 physically verified New York City addresses with
a clean lead or non-lead public-side material (15,154 lead, 9.18\%; 2,198 recorded galvanized
or unknown despite physical verification are excluded and counted). Features are coordinates
and building type, and nothing else: the MapPLUTO join is used by the estimators of
Section~\ref{sec:nyc} but not here, so every figure in this section rests on three columns
that any state's inventory would already contain. Customer-side fields are excluded because
some are themselves model-derived, and using them would leak the vendor model's output into a
supposedly independent baseline.

\begin{table}[t]\centering\small
\caption{Cross-validated discrimination as a function of how far apart training and test data
are held. A whole group falls in one fold.}
\label{tab:cv}
\begin{tabular}{lrrr}
\toprule
Cross-validation grouping & Groups & AUC, no ZIP feature & AUC, with ZIP feature \\
\midrule
ZIP code       & 200 & 0.720 & 0.638 \\
1\,km block    & 822 & 0.736 & 0.730 \\
2\,km block    & 249 & 0.727 & 0.705 \\
5\,km block    &  61 & 0.639 & 0.660 \\
10\,km block   &  21 & 0.645 & 0.637 \\
\bottomrule
\end{tabular}
\end{table}

Table~\ref{tab:cv} reports discrimination against train--test separation, a diagnostic we
consider mandatory for spatial prediction claims and did not find in this literature. The
honest summary is a range, 0.64--0.74, not a point. The right-hand column is a caution for
implementers: including ZIP as a feature while also grouping folds by ZIP depresses measured
AUC by 0.08, because the feature is dead at test time but still consumes splits. A pipeline
that does this and then refits on all data to produce a headline number reports an accuracy
that does not describe the model it deployed. We note the block-size column confounds edge
length with group count and read it as bounding, not decomposing, spatial leakage.

Two diagnostics matter more than the AUC, because the estimates above are sums of predicted
probabilities and AUC is invariant to whether those probabilities are on the right scale.
Out-of-fold, $\sum\hat p/\sum y = 0.875$ overall and 0.866 in the band the model-cleared
population occupies, so the estimator undercounts by roughly 13\% and the reported band is
conservative. A classifier trained to distinguish the model-cleared population from the
physically verified one reaches AUC 0.753 under spatial-block cross-validation: the two
populations are strongly separable on covariates, which is why the era join was necessary.

At 0.64--0.74 on public covariates alone this is a floor on the difficulty of the task, not a
ceiling on what a vendor model with work orders, permit history and assessor records could
achieve. That distinction must accompany the number. Some of the gap is cheap to close: adding
the MapPLUTO year built and built-form columns to the same ZIP-grouped design lifts AUC from
0.720 to 0.755, from one free municipal dataset, which is why we treat 0.74 as a floor rather
than a target. It also cuts against the outputs audited here. If coordinates and building type
alone separate lead from non-lead this well, an output with no variation at all is harder to
explain, not easier.

\section{Threats to validity}\label{sec:threats}

\paragraph{Conditional exchangeability is an assumption.}
Stated above with its bias direction. The estimator band is conditional on it; the screen and
the two dated facts are not.

\paragraph{Excavation and field inspection are not one instrument.}
We pool them but they differ: 10.13\% against 8.48\% lead on clean-label addresses
($\chi^2$, $p = 2.8\times10^{-30}$). Excavation records an unknown material on 2.60\% of its
addresses against field inspection's 0.01\%, and a method that sees less of the line returning
far fewer unknowns is the wrong way round. The guidance lists the two separately
\cite{nysdohguidance2025}, and field inspection may observe only the accessible portion, so
neither is ground truth in the laboratory sense and the pooled label is the weaker one.

Repeating the analysis with excavation as the only physical method leaves the results
standing. The screen is unchanged at 75 single-value localities and the same seven
contradicted; East Rochester's 557 verified lines are all excavations already; the pre-1940
physical lead rate moves from 12.93\% to 13.35\%, raising the pre-1940 implied count from
about 1,000 to about 1,040. The estimator band widens downward, to roughly 850--1,325, because
excavation alone supports fewer strata at the 20-address minimum.

\paragraph{\KO{} carries two readings.}
It appears in the dropdown beside named materials, and the guidance also permits it where the
material is not known but lead-ban records exist \cite{nysdohguidance2025}. We rely only on
what both readings share: it asserts a known negative lead status, and it is not one of the
three values available for recording an unknown.

\paragraph{Building age is a proxy.}
MapPLUTO year built describes the structure, not the service line. A line may have been
replaced after the building went up, which makes the proxy too pessimistic, and the inventory
records no replacement dates for these addresses. The reverse case matters more here: a
building may have been rebuilt over a line that was already in the ground, which makes the
proxy too optimistic. Half of the model-cleared population sits in post-1980 buildings, and
physical verification finds lead in post-2000 buildings at 2.98\% ($n = 13{,}683$), above the
1.45\% of the 1980--2000 bucket. That inversion is what old lines under new buildings look
like, and it is not a small part of the total: the post-2000 bucket contributes 318 of the
1,738 lines in the borough-blind era-stratified count, 18\%. Treating the post-2000 rate as
spurious and forcing it down to the 1980--2000 rate lowers the era-aware estimators by about
11\%, to roughly 1,140--1,300. The estimate is sensitive at that level to whether year built
is a valid proxy in the newest bucket.

MapPLUTO's year built is also heaped on round numbers: 44.9\% of joined addresses fall on a
multiple of ten and 71.0\% on a multiple of five. The era bin edges sit on the heap, and
because bins are left-closed a lot recorded as 1940 is counted post-1940, so the 7,782
pre-1940 figure is more likely an undercount than an overcount.

\paragraph{Scope.}
One state; the screen covers localities with at least 100 model-classified addresses, and the
detailed analysis is five boroughs, public side. The Buffalo-metro cluster, eighteen locality
values that are probably one utility relationship, is untouched.

\paragraph{What would falsify this.}
A per-address record linking a model prediction to a subsequent excavation showing the cleared
lines are lead-free at a rate consistent with zero; or documentation that the
\textsf{Statistical Analysis/Predictive Model} label is applied only after confirmation by
another means; or the confidence interval and verification plan the guidance asks systems to
supply. Any of these could be produced by a utility or vendor in an afternoon.

\section{Related work}

Independent validation of a lead service line prediction model against excavation exists once,
for Flint \cite{goovaerts2023}. Vendor-adjacent work reports higher figures on private test
sets \cite{deheer2023}, and a physical-sensing study on synthetic data reports 99.9\%
\cite{acsest2026}; the published range runs from roughly 73\% to 99.9\% with no shared
benchmark, which is itself evidence these numbers are not comparable. Smart et al.
\cite{bennington2023} document one community's identification programme in detail and report
what its methods cost and found, which is the kind of account this literature has few of. A
qualitative review \cite{review2021} confirms no quantitative independent benchmark exists.
Nigra et al.
\cite{nigra2023} model lead service line predictors in New York City on an older,
records-based, pre-LCRR dataset and independently establish the MapPLUTO join used here; their
concern is environmental-justice patterning, and their data lacks the verification-method
field this audit depends on. The New York State Comptroller audited this programme through
June 2025, including whether the inventory was completed accurately \cite{osc2026}; that
report does not examine the predictive-model pathway. We are aware of no independent audit of
an LCRR-era commercial prediction model.

\section{Discussion}

We have not shown that predictive models for service lines cannot work; Flint suggests they
work poorly and our own baseline suggests the task is tractable but hard. We have shown that
across half the New York systems using one, the recorded output carries no variation, that
seven of those are contradicted by their own crews, and that in the largest case the filing
does not meet the conditions the state's own guidance attaches to the value used.

Three things follow without agreeing with our estimate.

First, \textbf{inventories should record the model's uncertainty, not its argmax}. New York
already asks systems for a confidence interval \cite{nysdohguidance2025} and its template
already contains the hedge \CBL{}, used 15,285 times elsewhere in the state. An inventory that
provides for uncertainty and receives none from a 43,215-address deployment is recording less
than it was designed to.

Second, \textbf{zero output variance is a red flag a regulator can check with one query}, and
the two-stage form matters: flag on variance, confirm against the system's own verification.
We ran it over the whole state in seconds. It found the largest deployment in New York and an
unrelated village, and it correctly declined to flag 68 systems whose single value is
consistent with what their crews find.

Third, \textbf{this audit is possible only because New York publishes the basis of
classification per address}. Most states do not. That one column is what makes the file
checkable, and requiring it nationally would cost nothing and make every vendor model in the
country auditable by anyone.

\section{Reproducibility}\label{sec:repro}

All inputs are public, and the code, the manuscript source and the recorded checksums are at
\url{https://github.com/msarmadsohail/predictive-lead-service-line-audit}. The state inventory
is retrievable from the Socrata endpoint in
\cite{nyslsl2026}; because the live dataset moves, both snapshots are pinned there by SHA-256,
and the 2025 snapshot is available from the Internet Archive. MapPLUTO is retrieved by an
included script that verifies the returned row count against the API's own count before
writing. The analysis is a chain of scripts shipped as ancillary files with this submission:
a one-pass extract of each snapshot, a canonical loader carrying the trap handling in
Section~\ref{sec:data}, the MapPLUTO fetch and spatial join, the statewide screen, the
snapshot comparison, the estimators, the baseline, and generators for every figure and number.
Each figure and table in this paper is produced from a JSON file of measured values written by
those scripts.

Three conventions are choices rather than consequences, so we state them. Address keys are
\mbox{locality}--\mbox{street}--\mbox{ZIP}, 98.97\% unique at T0. Where a key genuinely
repeats, at addresses carrying more than one service line, the New York City loader keeps the
first row in file order and the snapshot comparison keeps the row with the higher-sorting
material. Neither is principled; both are reproducible from the pinned captures. In New York
City the choice affects 607 rows of 817,982 and moves the city lead count by 0.06\%, and the
model bucket is \KO{} on every row under either rule. Second, the rest-of-state comparison in
Section~\ref{sec:nyc} deduplicates after selecting the method while the screen deduplicates
before, which is why the two report 176,888 and 176,462 addresses for what reads as the same
population. Third, era bins are left-closed, which matters given the year-built heaping noted
in Section~\ref{sec:threats}.

\bibliographystyle{plain}
\bibliography{refs}

\end{document}